\documentclass[conference]{IEEEtran}
  \IEEEoverridecommandlockouts

  \usepackage{cite}
  \usepackage{amsmath,amssymb,amsfonts}
  \usepackage{graphicx}
  \usepackage{textcomp}
  \usepackage{xcolor}
  \usepackage{booktabs}
  \usepackage{multirow}
  \usepackage{url}
  \usepackage{tikz}
  \usepackage[absolute,overlay]{textpos}

  \def\BibTeX{{\rm B\kern-.05em{\sc i\kern-.025em b}\kern-.08em
      T\kern-.1667em\lower.7ex\hbox{E}\kern-.125emX}}

  \newcommand{\qgate}[3]{%
    \draw[fill=white] (#1-0.34,#2-0.30) rectangle (#1+0.34,#2+0.30);
    \node at (#1,#2) {\small $#3$};}

\begin{document}

  \title{Quantum Feature Engineering for Credit Default Prediction:
  When and Why IQP Circuits Help Linear Classifiers}

\author{
\IEEEauthorblockN{Menachem Finkelstein\IEEEauthorrefmark{1},
Diana Legziel Levy\IEEEauthorrefmark{2}, Zohar Yakhini\IEEEauthorrefmark{3},
and Sarel Cohen\IEEEauthorrefmark{4}}
\IEEEauthorblockA{\textit{Reichman University}, Israel \\
\IEEEauthorrefmark{1}mnchmf@gmail.com \quad
\IEEEauthorrefmark{2}legziel.diana@post.runi.ac.il \\
\IEEEauthorrefmark{3}zohar.yakhini@gmail.com \quad
\IEEEauthorrefmark{4}sarel.cohen@runi.ac.il}
}

\maketitle

\enlargethispage{-0.65in}

\begin{textblock*}{\textwidth}(0.75in,10.1in)
\scriptsize
\copyright~2026 IEEE. Personal use of this material is permitted.
Permission from IEEE must be obtained for all other uses, including
reprinting/republishing this material for advertising or promotional purposes,
collecting new collected works for resale or redistribution to servers or lists,
or reuse of any copyrighted component of this work in other works.
\end{textblock*}

  \begin{abstract}
  Credit default prediction is a tabular classification problem in which
  modest gains in F\textsubscript{1} translate directly into reduced
  financial exposure.  We ask whether \emph{Instantaneous Quantum
  Polynomial-time} (IQP) circuits can produce features that improve a
  classifier over both its raw classical baseline and Kernel PCA---the
  strongest unsupervised classical non-linear alternative---at an equal
  feature budget.  The dataset provides $23$ financial attributes per
  client; for an $n$-qubit circuit we select $n$ of them, encode each as a
  rotation angle, and read $2n$ expectation values back out as new
  features.  The motivation for using a quantum circuit is computational:
  an $n$-qubit IQP circuit runs in constant depth and encodes feature
  correlations in a $2^{n}$-dimensional Hilbert space, whereas classical
  simulation of its exact output statistics scales exponentially in $n$.
  Using the UCI \emph{Default of Credit Card Clients} dataset and five-fold
  cross-validation, we find that appending $16$ IQP features ($n=8$ qubits)
  to a Logistic Regression model raises F\textsubscript{1} from
  $0.462$ to $0.517$ ($+0.055$, $p<0.0001$).  Kernel PCA, the next-best
  method, reaches only $0.493$ at the same feature count; the gap survives
  Benjamini--Hochberg correction across $12$ tests ($p=0.00007$).  No other
  classifier---Random Forest, SVM, XGBoost, or k-NN---benefits, which
  points to a linear-expressivity mechanism rather than a generic
  improvement.  We also show that \emph{how} the $8$ input features are
  chosen matters: Random Forest importance-guided selection reaches
  F\textsubscript{1}$=0.523$, while encoding maximally uncorrelated
  features drops it to $0.496$, demonstrating that the circuit amplifies
  informative structure rather than creating it from scratch.
  \end{abstract}


  \section{Introduction}

  Credit card default prediction sits at the intersection of class
  imbalance, non-linear feature interactions, and regulatory pressure to
  use interpretable models.  The UCI \emph{Default of Credit Card Clients}
  dataset describes each client with $23$ heterogeneous features---credit
  limits, demographics, and six months of payment history---and only
  $\sim\!22.6\%$ of clients default, so it is a realistic stress test for
  any feature-engineering strategy.

  Logistic Regression is the workhorse of credit scoring, but its
  limitation is structural: it fits a single hyperplane in the original
  $23$-dimensional feature space, so payment-history variables that
  interact \emph{multiplicatively} cannot be separated by a straight
  decision boundary.  Quantum feature maps address this by embedding the
  data into a high-dimensional Hilbert space in which those interactions
  are represented implicitly through entanglement
  \cite{havlicek2019, schuld2019}.  Concretely, a parametric circuit maps a
  real input vector into a state in a Hilbert space of dimension $2^{n}$,
  and the expectation values of local observables on the output state become
  new features that encode non-linear combinations of the inputs.  For IQP
  circuits specifically, the argument for eventually using real quantum
  hardware is not only expressivity but \emph{cost}: the circuit runs in
  constant depth for any number of qubits $n$, whereas classically
  simulating its exact output distribution requires time exponential in
  $n$ \cite{shepherd2009, bremner2016, coyle2020}.

  This paper studies the question empirically and on a classical simulator;
  that is, we ask whether IQP-derived features, \emph{however they are
  computed}, improve a downstream linear classifier over the best classical
  alternative.  Whether quantum hardware could compute these same features
  more cheaply than classical simulation is a separate, complexity-theoretic
  matter that we return to in Section~\ref{sec:limitations}.  Two concrete
  questions drive the work.  First, do IQP features improve over the raw
  $23$-feature baseline when appended to a Logistic Regression model?
  Second, do they beat Kernel PCA (KPCA) at an equal feature budget?  KPCA
  is the appropriate \emph{target of comparison} because it is, like the
  quantum map, unsupervised, non-linear, and produces abstract features
  without ever looking at the class label.  Beating KPCA would mean the
  quantum map captures structure that the best classical unsupervised
  method misses.

  Prior work on quantum machine learning for finance tends to test on
  small datasets, skip multiple-comparison corrections, or compare quantum
  against a weak baseline rather than the best available classical
  alternative \cite{liu2021}.  We address all three issues.  Our
  contributions are: (i) a nine-way equal-budget comparison of quantum and
  classical feature-extraction methods on a standard financial dataset;
  (ii) evidence that the quantum advantage is specific to linear
  classifiers and absent for Random Forest, SVM, XGBoost, and k-NN;
  (iii) false-discovery-rate (FDR) corrected significance testing over $12$
  planned comparisons;
  and (iv) a study of five feature-selection strategies showing that
  RF-importance-guided selection outperforms arbitrary choice and that
  encoding uncorrelated features actively hurts performance.

  \section{Background and Related Work}

  \subsection{Quantum Feature Maps}

  A quantum feature map $\phi:\mathbb{R}^{d}\rightarrow\mathcal{H}$ embeds
  a classical vector $\mathbf{x}$ into a quantum state
  $|\phi(\mathbf{x})\rangle$ living in a Hilbert space $\mathcal{H}$ of
  dimension $2^{n}$, where $n$ is the number of qubits.  To turn that state
  back into usable numbers we measure a fixed set of $K$ Hermitian
  \emph{observables} $\{O_{k}\}_{k=1}^{K}$ (in this work the per-qubit Pauli
  $X$ and $Y$ operators, so $K=2n$); each one yields a scalar feature
  \begin{equation}
    f_{k}(\mathbf{x}) =
    \langle\phi(\mathbf{x})\,|\,O_{k}\,|\,\phi(\mathbf{x})\rangle,
    \qquad k = 1,\dots,K .
  \end{equation}
  The index $k$ simply enumerates the observable (equivalently, the output
  feature).  The corresponding kernel
  $\kappa(\mathbf{x},\mathbf{x}') =
  |\langle\phi(\mathbf{x})|\phi(\mathbf{x}')\rangle|^{2}$
  can represent decision boundaries that standard kernels express only with
  a very large number of terms---for example boundaries that depend on the
  \emph{parity} or simultaneous sign of several features (XOR-like
  interactions), which a polynomial or RBF kernel approximates poorly at low
  order \cite{havlicek2019}.

  \subsection{IQP Circuits}

  IQP (\emph{Instantaneous Quantum Polynomial-time}) circuits interleave
  Hadamard layers with diagonal commuting unitaries.  Their output
  distributions are believed to be hard to sample classically
  \cite{shepherd2009, bremner2016}, and on quantum hardware their depth is
  \emph{independent} of the number of qubits, which makes them attractive
  for near-term devices.  Coyle et al.\ \cite{coyle2020} propose IQP
  circuits as practical feature extractors precisely because expectation
  values are easy to estimate by sampling even when exact simulation is
  hard.  The general feature-map formalism of the previous subsection
  applies directly to IQP; the next section makes the specific circuit,
  its inputs, and its outputs fully concrete.

  \subsection{Credit Risk Modeling}

  Logistic Regression has dominated credit scoring for decades because
  regulators can inspect its coefficients and auditors can reconstruct
  individual decisions \cite{siddiqi2012}.  Kernel SVMs and gradient
  boosting have shown competitive accuracy \cite{dastile2020}, but remain
  black boxes that are difficult to certify under frameworks such as the
  EU's model-risk guidelines.  Quantum-enhanced approaches for credit
  modelling are nascent; most published results use toy datasets or do not
  compare against strong classical baselines.

  \section{Methodology}

  \subsection{Dataset}

  We use the UCI \emph{Default of Credit Card Clients} dataset
  (OpenML \#42477) \cite{yeh2009}, using all $N=30{,}000$ clients for
  every experiment.  The original input
  is a vector of $23$ features per client (credit limits, demographic
  variables, and six monthly payment-history records); the positive
  (default) rate is $22.6\%$, creating a mild class imbalance.  All
  experiments use five-fold stratified cross-validation, and reported
  metrics are the mean $\pm$ standard deviation over folds.

  \subsection{IQP Quantum Feature Extraction}
  \label{sec:iqp}

\begin{figure}[t]
  \centering
  \begin{tikzpicture}[x=1.0cm, y=0.85cm]
    \foreach \q/\lab in {0/1,1/2,2/3,4/7,5/8}{
      \node[anchor=east] at (-0.12,-\q) {\small $|0\rangle_{\lab}$};
      \draw (0,-\q) -- (6.90,-\q);
    }
    \node at (-0.42,-3) {\small $\vdots$};
    \foreach \q in {0,1,2,4,5}{\qgate{0.8}{-\q}{H}}
    \foreach \q/\lab in {0/1,1/2,2/3,4/7,5/8}{
      \draw[fill=white] (2.1-0.62,-\q-0.30) rectangle (2.1+0.62,-\q+0.30);
      \node at (2.1,-\q) {\scriptsize $R_Z(\theta_{\lab})$};}
    \foreach \x/\a/\b in {3.30/0/1, 3.85/1/2, 4.95/4/5}{
      \draw[thick] (\x,-\a) -- (\x,-\b);
      \fill (\x,-\a) circle (1.9pt);
      \fill (\x,-\b) circle (1.9pt);}
    \node at (4.40,-3) {\small $\cdots$};
    \draw[thick,dashed] (5.60,0) -- (5.60,-5);
    \fill (5.60,0) circle (1.9pt); \fill (5.60,-5) circle (1.9pt);
    \node[anchor=north,align=center] at (5.60,-5.42)
      {\scriptsize $\mathrm{ZZ}_{8,1}$ (wrap-around)};
    \node[anchor=south] at (4.1,0.18) {\scriptsize $\mathrm{ZZ}$ ring layer};
    \foreach \q in {0,1,2,4,5}{%
      \draw[fill=black!6] (6.25,-\q-0.30) rectangle (6.90,-\q+0.30);
      \node at (6.57,-\q) {\scriptsize $X,Y$};}
  \end{tikzpicture}
  \caption{The $8$-qubit IQP feature extractor (five of the eight wires
    shown; qubits $4$--$6$ and the couplings $(3,4)\dots(6,7)$ are elided).
    Left to right: a Hadamard superposition layer, an $R_{Z}(\theta_{i})$
    data-encoding layer, and an \textsc{IsingZZ} entanglement layer.  Each
    qubit is then measured in the Pauli $X$ and $Y$ bases, producing
    $2n=16$ output features; that transverse read-out plays the role of the
    canonical IQP form's closing basis change, and is why
    $\langle Z\rangle$ is identically zero and excluded.
    \emph{Topology:} the ZZ couplings form a \emph{ring} on the $8$
    qubits---exactly the $8$ edges $(1,2),(2,3),\dots,(7,8),(8,1)$, each
    applied at its own time step.  The dashed line is the wrap-around edge
    $(8,1)$ that closes the ring; it couples qubit $8$ to qubit $1$ only,
    crossing the intervening wires without acting on them.}
  \label{fig:circuit}
  \end{figure}
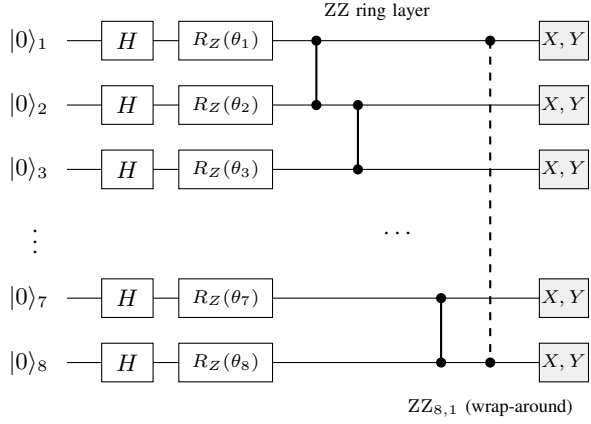

  This section is the core of the method, so we describe it end to end:
  what enters the circuit, what the circuit does, and what comes out.
  Figure~\ref{fig:circuit} summarises the three-layer circuit, and the
  text below follows the same order.

  \paragraph{What enters the circuit (the input)}
  The circuit has $n$ qubits; throughout the main experiments $n=8$.
  Because $8$ qubits can accept only $8$ angles, we first \emph{select} $8$
  of the $23$ classical features (the selection strategy is itself studied
  in Section~\ref{sec:featsel}).  Each selected feature is standardised with
  a \texttt{QuantileTransformer} (mapping it to a normal distribution),
  clipped to $[-3\sigma,3\sigma]$, and converted to an encoding angle
  \begin{equation}
    \theta_{i} = x_{i}\cdot \pi/3 , \qquad i = 1,\dots,n ,
  \end{equation}
  which places every angle in $(-\pi,\pi)$ and so uses the full dynamic
  range of the rotation gates.  The vector
  $\boldsymbol{\theta}=(\theta_{1},\dots,\theta_{8})$ is the actual input to
  the circuit: one angle per qubit.

  \paragraph{What the circuit does (the three layers)}
  The $8$-qubit IQP unitary $U_{\mathrm{IQP}}(\boldsymbol{\theta})$ is built
  from three layers, shown left to right in Fig.~\ref{fig:circuit}:

  \begin{enumerate}
    \item \textbf{Superposition layer.}  A Hadamard gate $H$ is applied to
      every qubit, taking the all-zero state into a uniform superposition
      over all $2^{8}=256$ computational basis states:
      \begin{equation}
        |0\rangle^{\otimes 8} \xrightarrow{H^{\otimes 8}}
        |+\rangle^{\otimes 8} =
        \frac{1}{\sqrt{2^{8}}}\sum_{x\in\{0,1\}^{8}}|x\rangle .
      \end{equation}

    \item \textbf{Data encoding.}  A single-qubit $R_{Z}(\theta_{i})$
      rotation imprints feature $\theta_{i}$ as a phase on qubit $i$:
      \begin{equation}
        R_{Z}(\theta_{i}) = \begin{pmatrix} e^{-i\theta_{i}/2} & 0 \\
                                        0 & e^{i\theta_{i}/2} \end{pmatrix}.
      \end{equation}
      The input is now stored in the \emph{phase} of the state, not its
      amplitude.

    \item \textbf{Entanglement layer.}  Two-qubit \textsc{IsingZZ} gates are
      applied on a ring (qubit $i$ coupled to qubit $i{+}1 \bmod 8$):
      \begin{equation}
        \mathrm{ZZ}_{ij}(\phi_{ij}) =
        e^{-i\phi_{ij}\,Z_{i}\otimes Z_{j}/2},
        \quad \phi_{ij} = \tfrac{1}{2}(\theta_{i}+\theta_{j}).
      \end{equation}
      This layer is the source of the quantum advantage: it
      \emph{entangles} neighbouring qubits, encoding pairwise (and, through
      the ring, higher-order) interactions between features in a way that
      has no efficient classical description as $n$ grows.

  \end{enumerate}

  Composing the three layers gives
  \begin{equation}
    U_{\mathrm{IQP}}(\boldsymbol{\theta}) =
      \Bigl(\textstyle\prod_{\langle i,j\rangle}
            \mathrm{ZZ}_{ij}(\phi_{ij})\Bigr)
      \Bigl(\textstyle\prod_{i} R_{Z}(\theta_{i})\Bigr)
      H^{\otimes 8},
  \end{equation}
  a single Hadamard layer followed by a block of commuting diagonal
  gates---the standard IQP feature map.  Reading the transverse observables
  $X$ and $Y$ rather than $Z$ folds the canonical IQP form's closing basis
  change into the measurement itself.  It also explains why the $Z$
  expectations vanish: writing $D$ for the diagonal block, $D$ commutes with
  $Z_{i}$, so $\langle Z_{i}\rangle = \langle+|^{\otimes 8}Z_{i}
  |+\rangle^{\otimes 8} = 0$ for every qubit and every input.

  \paragraph{What comes out (the output features)}
  After applying $U_{\mathrm{IQP}}$ we measure, for each qubit $i$, the
  expectation values of the Pauli $X$ and $Y$ observables:
  \begin{equation}
    f_{X,i} = \langle\psi|X_{i}|\psi\rangle, \qquad
    f_{Y,i} = \langle\psi|Y_{i}|\psi\rangle ,
  \end{equation}
  where $|\psi\rangle = U_{\mathrm{IQP}}(\boldsymbol{\theta})|0\rangle^{\otimes 8}$
  is the state the circuit prepares.
  This yields $8\times 2 = 16$ quantum features per client.  (As shown
  above, the Pauli $Z$ expectations are identically zero for this circuit
  and are therefore excluded, which is why we read $2n$ rather than $3n$
  features.)  Each
  output feature is a non-linear function of \emph{all} the input angles
  $\theta_{i}$ that reflects inter-qubit entanglement; it cannot be factored
  into independent per-qubit contributions.

  \paragraph{Combined representation}
  The $16$ quantum features are concatenated with the original $23$
  classical features, giving a $39$-dimensional input vector that is then
  fed to the downstream classifier.  In other words, the quantum circuit is
  used as a feature \emph{augmenter}, not a replacement.

  \paragraph{Encoding variants}
  We test two encodings.  The IQP encoding above uses the ZZ layer to couple
  qubit phases.  As a simpler control, \emph{Angle encoding} reduces the
  circuit to a single $R_{Y}(\theta_{i})$ rotation per qubit---no Hadamard
  layer and no entangling gates---read out in the same $X$ and $Y$ bases,
  producing $16$ features with \emph{no} inter-qubit coupling.  Both encodings are evaluated across all five classifiers,
  yielding $10$ encoding--classifier pairs; each pair is also compared
  directly against KPCA, adding $2$ more tests.  This gives the family size
  $m=12$ used in the multiple-comparison correction
  (Section~\ref{sec:fdr}).

  \subsection{Why This Circuit, and Why $n=8$?}
  \label{sec:rationale}

  \paragraph{Why $n=8$ qubits}
  The qubit count was fixed \emph{a priori} by the experimental design and
  was not selected on performance.  An $n$-qubit circuit emits $2n$
  features---Pauli $X$ and $Y$ on each qubit---so $n=8$ emits exactly $16$,
  which is precisely the feature budget every classical comparator is given
  in Table~\ref{tab:logreg_comparison}.  Fixing $n$ this way is what makes
  the nine-way comparison like-for-like; choosing $n$ to maximise
  F\textsubscript{1} instead would have made the budgets unequal and would
  have spent a degree of freedom that the FDR family in
  Section~\ref{sec:fdr} does not account for.

  \paragraph{Why this circuit}
  The three layers of Section~\ref{sec:iqp} are not interchangeable design
  choices; each is forced by a requirement.  Phase encoding is dictated by
  the IQP form itself, whose data-dependent part must be diagonal in the
  computational basis, and it is also the cheapest loading available---one
  gate per feature at depth $1$, against amplitude encoding, whose loading
  circuits are exponentially deep in $n$.  The single Hadamard layer is what
  makes that phase information matter, since it spreads the state over all
  $2^{n}$ basis states before the phases are written; the transverse
  read-out then supplies the interference that turns those phases into
  measurable numbers, at no gate cost.  The one genuinely
  free choice is the coupling topology, and we take a \emph{ring} over
  all-to-all because a ring has $n$ edges and is $2$-edge-colourable,
  giving $O(n)$ gates at constant depth, whereas all-to-all needs
  $n(n{-}1)/2$ gates at a depth that grows with $n$.  Constant depth is the
  entire near-term argument for IQP, so we do not trade it away, and a ring
  is in any case the connectivity most readily available on near-term
  hardware.  We make no novelty claim for the encoding itself---it is the
  standard IQP feature map of Havl\'i\v{c}ek et al.\ \cite{havlicek2019}
  and Coyle et al.\ \cite{coyle2020}.  What this paper contributes is the
  controlled, equal-budget, FDR-corrected account of what such a map is and
  is not worth on a real tabular problem.

  \paragraph{Depth and gate count}
  For even $n$, the circuit of Section~\ref{sec:iqp} uses $n$ Hadamard,
  $n$ $R_{Z}$ and $n$ \textsc{IsingZZ} gates---$3n$ gates in all, of which
  $n$ are two-qubit---at depth $4$: one Hadamard layer, one $R_{Z}$ layer,
  and two ZZ sub-layers (an even ring is $2$-edge-colourable, so its
  couplings execute in two rounds).  At $n=8$ that is $8$ Hadamard, $8$
  $R_{Z}$ and $8$ ZZ gates, i.e.\ $24$ gates at depth $4$.  Compiled to a
  CNOT-and-rotation gate set, where
  $\mathrm{ZZ}(\phi)=\mathrm{CNOT}\cdot R_{Z}(\phi)\cdot\mathrm{CNOT}$, the
  same circuit becomes $8$ Hadamard, $16$ $R_{Z}$ and $16$ CNOT gates at
  depth $8$.  The property that matters is that both depths are
  \emph{independent of $n$}: widening the circuit adds gates but no layers.
  The circuit also has no trainable parameters---its angles are data, not
  weights---so it contributes nothing to the parameter count of the model
  that consumes its output.

  \subsection{Classical Comparison Methods}

  Each classical method adds exactly $16$ features, matching the quantum
  budget ($39$ total).  We include a noise baseline ($16$ random Gaussian
  features) as a sanity check against dimensionality inflation, and a family
  of \emph{linear} transformations---PCA, SVD, ICA, random projection, and
  feature agglomeration---to establish whether \emph{any} unsupervised
  transformation helps a linear classifier at all.  Polynomial degree-$2$
  terms test explicit pairwise interactions.  Kernel PCA with an RBF kernel
  (bandwidth $\gamma=1/(d\cdot\mathrm{Var}(\mathbf{x}))$, the
  \texttt{sklearn} default) is our \emph{primary} classical comparator:
  like the quantum approach it is unsupervised, non-linear, and produces
  abstract features without access to the class label.  Methods that use
  supervised feature selection (e.g.\ polynomial expansion followed by
  \texttt{SelectKBest}) have an inherent advantage and are reported for
  completeness only.

  Two further controls hold by construction rather than by tuning.  Every
  augmented arm in Table~\ref{tab:logreg_comparison} presents the
  classifier with the same $39$ inputs, hence the same number of fitted
  parameters ($39$ coefficients plus an intercept), and every arm is
  trained on the same $30{,}000$ clients under the same fold partition;
  only the \emph{content} of the $16$ appended columns differs.  Because
  the quantum circuit has no trained weights of its own, matching parameter
  count and training-set size across quantum and classical arms is not a
  separate experiment here---it is a property of the equal-budget design.

  \subsection{Classifiers}

  We evaluate five classifiers: Logistic Regression (L2, $C=1.0$, balanced
  weights), Random Forest ($300$ trees, balanced), SVM (RBF kernel),
  XGBoost ($100$ trees, $\mathit{lr}=0.1$), and k-NN ($k=5$,
  distance-weighted).  All use \texttt{sklearn} defaults except where
  stated.

  \subsection{Statistical Testing}

  Significance is assessed with paired $t$-tests across the five CV folds
  ($\alpha=0.05$).  To control for multiple comparisons we apply the
  Benjamini--Hochberg (BH) FDR procedure \cite{benjamini1995} over a family
  of $12$ planned comparisons.

  \section{Experimental Results}

  \subsection{Effect Across Classifiers}

  Table~\ref{tab:all_classifiers} reports each classifier with and without
  the $16$ appended IQP features; values are mean $\pm$ standard deviation
  over the five folds.  The pattern is stark: Logistic Regression gains
  $0.055$ F\textsubscript{1} and $8.4$ percentage points of accuracy (both
  $p<0.0001$), while every non-linear classifier is statistically
  unchanged---Random Forest, SVM, XGBoost, and k-NN all return $p>0.05$ on
  every metric, and several change by less than one standard deviation.

  The reason is structural.  A linear classifier cannot form non-linear
  decision boundaries on its own, so the quantum features supply the
  non-linearity externally.  Ensemble and kernel methods already construct
  such boundaries internally---through tree splits, an RBF kernel, or
  boosting residuals---so the same features carry redundant information for
  them.  This also rules out a generic dimensionality-inflation effect:
  if merely adding $16$ extra inputs helped, we would see gains across all
  five classifiers, which we do not.

  \begin{table}[t]
  \centering
  \caption{Performance with and without $16$ IQP features, per classifier.
    ``$+$Quantum'' uses the $8$-qubit IQP encoding.  Values are
    mean\,$\pm$\,std over $5$ folds.  Only Logistic Regression improves
    significantly.}
  \label{tab:all_classifiers}
  \setlength{\tabcolsep}{3pt}
  \footnotesize
  \begin{tabular}{llccc}
  \toprule
  \textbf{Classifier} & \textbf{Method} & \textbf{F\textsubscript{1}} & \textbf{Acc.\,(\%)} & \textbf{AUC} \\
  \midrule
  \multirow{2}{*}{Log.\ Reg.}
    & Classical       & $0.462${\scriptsize$\pm.014$} & $67.5${\scriptsize$\pm.3$} & $0.708${\scriptsize$\pm.014$} \\
    & \textbf{$+$Quantum} & $\mathbf{0.517}${\scriptsize$\pm.017$} & $\mathbf{75.9}${\scriptsize$\pm.9$} & $\mathbf{0.744}${\scriptsize$\pm.014$} \\
  \midrule
  \multirow{2}{*}{Rand.\ For.}
    & Classical       & $0.500${\scriptsize$\pm.026$} & $79.9${\scriptsize$\pm.5$} & $0.753${\scriptsize$\pm.010$} \\
    & $+$Quantum  & $0.494${\scriptsize$\pm.024$} & $79.9${\scriptsize$\pm.5$} & $0.755${\scriptsize$\pm.013$} \\
  \midrule
  \multirow{2}{*}{SVM (RBF)}
    & Classical       & $0.517${\scriptsize$\pm.017$} & $76.1${\scriptsize$\pm.8$} & $0.737${\scriptsize$\pm.017$} \\
    & $+$Quantum  & $0.519${\scriptsize$\pm.018$} & $76.1${\scriptsize$\pm1.1$} & $0.745${\scriptsize$\pm.016$} \\
  \midrule
  \multirow{2}{*}{XGBoost}
    & Classical       & $0.517${\scriptsize$\pm.017$} & $75.5${\scriptsize$\pm1.0$} & $0.748${\scriptsize$\pm.007$} \\
    & $+$Quantum  & $0.509${\scriptsize$\pm.027$} & $76.4${\scriptsize$\pm1.7$} & $0.739${\scriptsize$\pm.017$} \\
  \midrule
  \multirow{2}{*}{k-NN}
    & Classical       & $0.396${\scriptsize$\pm.009$} & $77.9${\scriptsize$\pm.3$} & $0.683${\scriptsize$\pm.014$} \\
    & $+$Quantum  & $0.393${\scriptsize$\pm.011$} & $77.8${\scriptsize$\pm.4$} & $0.681${\scriptsize$\pm.015$} \\
  \bottomrule
  \end{tabular}
  \end{table}

  \subsection{Nine-Way Comparison at Equal Feature Budget}

  Table~\ref{tab:logreg_comparison} ranks all nine feature-engineering
  methods at $39$ features each for Logistic Regression; in every case the
  \emph{same} $16$ extra features are computed once and reused inside each
  cross-validation fold (no per-fold refitting of the transform on test
  data).  Quantum features place first.  The five linear
  transformations---PCA, SVD, ICA, random projection, and feature
  agglomeration---all leave F\textsubscript{1} essentially unchanged
  ($\Delta\le 0.001$, $p>0.37$).  This is expected rather than surprising:
  a \emph{linear} transformation of features fed to a \emph{linear} model
  stays inside the same hypothesis class, so it cannot change the set of
  achievable decision boundaries---the linear methods serve as a control
  that confirms only genuinely non-linear maps can help here.  Polynomial
  degree-$2$ terms add $0.008$ F\textsubscript{1}, a real but modest gain.
  KPCA does substantially better, adding $0.031$ ($p=0.0001$), confirming
  that non-linearity is what matters.  Quantum features add $0.055$, roughly
  $1.8\times$ the KPCA gain ($p=0.00007$ for the quantum--KPCA gap).  The
  random-noise baseline is the one method that \emph{hurts}
  F\textsubscript{1} slightly, ruling out any explanation based purely on
  the increase in dimensionality.

  \begin{table}[t]
  \centering
  \caption{Logistic Regression with $16$ augmented features ($39$ total).
    All methods are unsupervised except Polynomial\textsubscript{16}
    (see text).  \textbf{Bold} marks the best result in the table.
    Primary metric is F\textsubscript{1}; $p$-values are from a paired
    $t$-test vs.\ the baseline.}
  \label{tab:logreg_comparison}
  \setlength{\tabcolsep}{4pt}
  \begin{tabular}{lcccc}
  \toprule
  \textbf{Method} & \textbf{F\textsubscript{1}} & $\Delta$ F\textsubscript{1} & \textbf{Acc.} & $p$-value \\
  \midrule
  Baseline (23)        & 0.462         & ---            & 67.5\,\%      & ---      \\
  Noise control        & 0.456         & $-$0.006       & 67.1\,\%      & 0.24     \\
  PCA\textsubscript{16} & 0.463        & $+$0.001       & 67.5\,\%      & 0.68     \\
  SVD\textsubscript{16} & 0.463        & $+$0.001       & 67.5\,\%      & 0.68     \\
  ICA\textsubscript{16} & 0.462        & $+$0.000       & 67.5\,\%      & 1.00     \\
  RandProj\textsubscript{16} & 0.463   & $+$0.001       & 67.5\,\%      & 0.73     \\
  FeatAgg\textsubscript{16} & 0.462    & $-$0.001       & 67.5\,\%      & 0.37     \\
  Poly\textsubscript{16} & 0.470       & $+$0.008       & 69.1\,\%      & 0.16     \\
  KPCA\textsubscript{16} & 0.493       & $+$0.031       & 73.1\,\%      & 0.0001   \\
  \textbf{Quantum\textsubscript{8q}} & \textbf{0.517} & \textbf{$+$0.055} & \textbf{75.9\,\%} & \textbf{$<0.0001$} \\
  \bottomrule
  \end{tabular}
  \end{table}

  \subsection{Multiple-Comparison Correction}
  \label{sec:fdr}

  With $12$ pre-specified tests---$10$ encoding--classifier pairs plus two
  direct quantum-versus-KPCA comparisons---at $\alpha=0.05$ we would expect
  roughly one false positive by chance alone.  The BH procedure sorts those
  $12$ $p$-values in ascending order and compares the one at rank $k$
  against the \emph{critical value} $\alpha k/m$, reported as ``BH crit.''\
  in Table~\ref{tab:fdr}; a test survives when its $p$-value lies at or
  below that value.  The critical value is a threshold rather than a score:
  it grows with rank, so later ranks face a more permissive bar, and a test
  that fails to survive is simply one whose evidence cannot be told apart
  from the false positives expected in a family of this size.

  Applying the correction (Table~\ref{tab:fdr}) leaves three findings
  intact: IQP
  outperforms both the baseline and KPCA for Logistic Regression, and Angle
  encoding outperforms the baseline.  Angle encoding does \emph{not} survive
  when compared directly to KPCA ($p\approx 0.12$), making IQP the only
  encoding with a defensible claim to genuinely exceed the best classical
  unsupervised alternative.  That non-survival is informative rather than
  unfortunate.  Angle encoding is precisely the ablation that deletes the ZZ
  layer, so the fact that it fails to beat KPCA while IQP beats it separates
  the contribution of \emph{entanglement} from the mere act of appending
  $16$ circuit-derived columns.  All eight tests involving non-linear
  classifiers remain non-significant.

  \begin{table}[t]
  \centering
  \caption{Benjamini--Hochberg (BH) FDR correction over the family of
    $m=12$ planned tests ($\alpha=0.05$); primary metric
    F\textsubscript{1}.  Rows $1$--$4$ are Logistic Regression; ranks
    $5$--$12$ are the remaining eight tests, either encoding against the
    baseline on each of the four non-linear classifiers.  \textbf{Rank} $k$
    orders the $12$ $p$-values ascending, and \textbf{BH crit.}\ is the
    threshold $\alpha k/m$ that a $p$-value must not exceed in order to
    survive (Section~\ref{sec:fdr}).}
  \label{tab:fdr}
  \setlength{\tabcolsep}{3pt}
  \footnotesize
  \begin{tabular}{clccc}
  \toprule
  \textbf{Rank} & \textbf{Hypothesis} & $p$-value & \textbf{BH crit.} & \textbf{Survives?} \\
  \midrule
  1 & IQP vs.\ KPCA         & 0.00007        & 0.0042 & \checkmark \\
  2 & IQP vs.\ Baseline     & $<0.0001$      & 0.0083 & \checkmark \\
  3 & Angle vs.\ Baseline   & $<0.0001$      & 0.0125 & \checkmark \\
  4 & Angle vs.\ KPCA       & $\approx$0.12  & 0.0167 & $\times$ \\
  5--12 & Quantum vs.\ Baseline & $\approx$0.5   & ---    & $\times$ \\
  \bottomrule
  \end{tabular}
  \end{table}

  \subsection{Quantum Features as Supplement, Not Replacement}

  To test whether quantum features can stand alone, we ran three IQP
  circuits in parallel---$8$, $8$, and $7$ qubits respectively---so that
  every one of the $23$ input features was encoded in at least one circuit.
  Each qubit contributes a Pauli $X$ and Pauli $Y$ measurement, giving
  $8{\times}2 + 8{\times}2 + 7{\times}2 = 46$ quantum features with complete
  feature coverage and no classical input.  Accuracy was $69.4\%$ and
  F\textsubscript{1}$=0.485$, well below the $75.9\%$ and $0.517$ achieved
  when the same circuits are combined with the original $23$ features.  Full
  coverage does not rescue quantum-only prediction: the model needs the raw
  financial variables, not just their quantum projections.

  \subsection{Feature Selection Strategy Comparison}
  \label{sec:featsel}

  An $8$-qubit circuit encodes exactly $8$ of the $23$ available features, so
  the choice of which $8$ to use is a real design decision.
  Table~\ref{tab:feature_selection} compares five strategies.

  \begin{table}[t]
  \centering
  \caption{F\textsubscript{1} by feature-selection strategy (Logistic
    Regression, $5$-fold CV).  Strategies S1--S4 append quantum features to
    all $23$ classical features ($39$ total); S5 uses $3$ circuits $+$ $23$
    classical ($69$ total).  \textbf{Bold} = best F\textsubscript{1}.}
  \label{tab:feature_selection}
  \setlength{\tabcolsep}{3pt}
  \begin{tabular}{llccc}
  \toprule
  \textbf{Strategy} & \textbf{Selection method} & \textbf{F\textsubscript{1}} & \textbf{Acc.} & \textbf{AUC} \\
  \midrule
  Baseline (no quantum)        & ---                    & 0.462 & 67.5\,\% & 0.708 \\
  \midrule
  S1: First-8 (arbitrary)      & Features 1--8          & 0.517 & 75.9\,\% & 0.744 \\
  \textbf{S2: RF Importance}   & \textbf{Top-8 by Gini} & \textbf{0.523} & 75.3\,\% & \textbf{0.748} \\
  S3: Correlation-based        & Top-8 corr.\ w/ target & 0.516 & 75.5\,\% & --- \\
  S4: Diverse (uncorrelated)   & Greedy min-corr        & 0.496 & 72.6\,\% & --- \\
  S5: Multi-circuit $+$ orig.  & All 23 (3 circuits)    & 0.519 & 74.6\,\% & --- \\
  \bottomrule
  \end{tabular}
  \end{table}

  The ordering in Table~\ref{tab:feature_selection} is not random.  RF
  importance (S2) is a deterministic, data-driven procedure: a Random
  Forest is trained on the full $23$ features and features are ranked by
  mean decrease in Gini impurity.  The same $8$ features are selected every
  time on the same data, unlike the arbitrary first-$8$ (S1), which simply
  takes whatever order the dataset happens to use.  The selected features
  are primarily payment-history and credit-limit variables
  ($\{f_{5}, f_{6}, f_{7}, f_{0}, f_{11}, f_{12}, f_{17}, f_{18}\}$), and the
  result, F\textsubscript{1}$=0.523$, is the highest across all strategies---
  that is, it slightly exceeds even the strong $0.517$ obtained by the
  arbitrary first-$8$.  Correlation-based selection (S3) uses similar logic
  and scores $0.516$, close but not identical.  The arbitrary first-$8$ (S1)
  lands in between at $0.517$; it happens to include several moderately
  predictive features, which explains why it is competitive despite no
  deliberate selection.

  The most instructive result is the diverse strategy (S4), which greedily
  selects features with the lowest mutual correlation.  F\textsubscript{1}
  drops to $0.496$---below every other strategy and statistically worse than
  first-$8$ ($p<0.01$).  Maximally uncorrelated features spread the
  circuit's encoding capacity across the full variable space, but most of
  those variables carry little signal about default.  The IQP circuit does
  not generate predictive power; it amplifies interactions among
  already-informative inputs.  When those inputs are weak, the entanglement
  layer has nothing useful to couple.

  Covering all $23$ features with three separate $8$-qubit circuits (S5)
  raises the feature count to $69$ and yields F\textsubscript{1}$=0.519$,
  better than arbitrary first-$8$ but still below RF importance and at
  considerably greater cost.  In practice, selecting the top-$8$ features by
  any reasonable ranking method is both simpler and more effective.

  \section{Discussion}

  \subsection{How Large Is the Effect?}
  \label{sec:effectsize}

  Two differences in this paper are easy to conflate, and only one of them
  carries the result.  The claim is the \emph{headline} comparison of
  Table~\ref{tab:logreg_comparison}: appending $16$ IQP features moves
  Logistic Regression from F\textsubscript{1}$=0.462$ to $0.517$---a gain
  of $+0.055$, or $12\%$ in relative terms---together with the $+0.024$
  margin over Kernel PCA at an identical feature budget.  Both survive BH
  correction across the family of $12$ planned tests.  For scale, $+0.055$
  is about three times the fold-to-fold standard deviation of the augmented
  model ($\pm0.017$ in Table~\ref{tab:all_classifiers}), and it comes with
  $8.4$ percentage points of accuracy.

  The $0.517$-versus-$0.523$ difference in Section~\ref{sec:featsel} is a
  \emph{secondary} comparison---between two ways of choosing which $8$ of
  the $23$ features to encode---and it is not the paper's claim.  At
  $0.006$ it is roughly a third of that same fold standard deviation, so we
  do not read the top of Table~\ref{tab:feature_selection} as a reliable
  ordering.  What that section does establish is the much larger,
  one-directional contrast between informative and uninformative inputs:
  $0.523$ against $0.496$, with the drop significant at $p<0.01$.

  On practical significance we would rather be exact than generous.  A
  $+0.055$ F\textsubscript{1} gain is real but incremental: the kind of
  change a lender measures across a book of accounts, not one an analyst
  would notice on any single application, and we do not claim it justifies
  quantum hardware on its own.  The transferable result is the
  \emph{mechanism}---that a constant-depth quantum feature map can supply a
  linear model with non-linear structure the strongest classical
  unsupervised alternative does not, and that the benefit appears only where
  the classifier cannot build such structure itself.

  \subsection{Why Only Logistic Regression?}

  The quantum kernel induced by the IQP circuit,
  $\kappa(\mathbf{x},\mathbf{x}') =
  |\langle\phi(\mathbf{x})|\phi(\mathbf{x}')\rangle|^{2}$, defines a
  decision boundary in a high-dimensional Hilbert space.  For Logistic
  Regression---a linear model in feature space---appending quantum features
  is equivalent to lifting the input into a richer representation where a
  linear boundary can capture non-linear structure in the original space.
  Random Forest, SVM with an RBF kernel, and XGBoost already implement their
  own non-linear transformations through tree splits, the RBF kernel matrix,
  or boosting residuals.  The IQP features add no information they cannot
  already express, so the gain is zero.

  \subsection{Feature Importance of Quantum Features}

  In the $39$-feature logistic regression model, the four largest absolute
  coefficients all belong to quantum features: \texttt{Q\_X1} ($0.766$),
  \texttt{Q\_X5} ($0.615$), \texttt{Q\_Y1} ($0.551$), and \texttt{Q\_Y6}
  ($0.474$).  Quantum features hold $10$ of the top-$20$ coefficient
  positions despite making up only $41\%$ of the feature set---meaning the
  model relies on them more heavily than their share of the input would
  predict.  This is not an artefact of scaling or regularisation; it
  reflects that quantum features carry predictive information the classical
  variables, fed individually into a linear model, do not capture on their
  own.

  \subsection{Limitations}
  \label{sec:limitations}

  All circuits were run on a classical simulator (\texttt{PennyLane}
  \texttt{default.qubit}) \cite{bergholm2018}.  Our claims are therefore
  about the
  \emph{predictive value} of IQP-derived features, not about a realised
  quantum speed-up: at $n=8$ qubits the statistics are trivially simulable,
  and the complexity-theoretic hardness only becomes relevant at much larger
  $n$.  Gate noise and decoherence on real hardware would also alter the
  expectation values and likely narrow the gap over KPCA.  Whether the
  advantage survives on near-term devices is an open question that should be
  settled before drawing hardware-deployment conclusions.

  The cost is worth being concrete about.  Each feature is the expectation
  of a $\pm1$-valued observable, so two-decimal precision needs on the order
  of $10^{4}$ shots, and since the $2n$ observables fall into just two
  commuting settings (all-$X$ and all-$Y$), the full $30{,}000$-client
  design needs on the order of $6\times10^{8}$ circuit executions.  Against
  a $+0.024$ F\textsubscript{1} margin and a fold spread of $\pm0.017$, a
  hardware replication would have to hold its estimation error well below
  the signal---which is why we treat it as separate work rather than an
  extra row in Table~\ref{tab:logreg_comparison}.

  \section{Conclusion}

  IQP feature extraction provides a statistically robust
  F\textsubscript{1} improvement for Logistic Regression on credit default
  prediction: $+0.055$ over the raw baseline ($p<0.0001$) and $+0.024$ over
  the best unsupervised classical alternative, Kernel PCA ($p=0.00007$,
  FDR-corrected over $12$ tests).  The same features produce no benefit for
  Random Forest, SVM, XGBoost, or k-NN, which locates the mechanism firmly
  in the inability of linear models to construct non-linear boundaries on
  their own.

  Feature selection for the quantum circuit is not a detail to be glossed
  over.  Encoding the top-$8$ features by Random Forest importance is the
  most reliable strategy (F\textsubscript{1}$=0.523$), while encoding
  maximally uncorrelated features actively damages performance
  (F\textsubscript{1}$=0.496$).  The circuit amplifies interactions among
  informative inputs; it cannot compensate for encoding variables that carry
  little signal about the outcome.

  Future work should evaluate these circuits on real quantum hardware with
  error mitigation, investigate whether variational training of the circuit
  parameters can widen that margin over Kernel PCA, and examine the
  fairness properties of quantum-augmented credit scores under
  consumer-protection regulations.

  \section*{Acknowledgment}

  In accordance with IEEE policy on AI-generated content, the authors
  disclose that Claude Sonnet (Anthropic), accessed through the Cursor
  editor, was used as a coding assistant when implementing the experiment
  and analysis scripts.  All experimental design choices, parameter settings, and
  reported results were determined and verified by the authors, who take
  full responsibility for the content of this paper.

  \bibliographystyle{IEEEtran}
  \bibliography{refs}

  \end{document}